# Enhanced Dengue Outbreak Prediction in Tamilnadu using Meteorological and Entomological data

Dr. Varalakshmi M, VIT Vellore, India, Dr. Daphne Lopez, VIT Vellore, India

Sponsored by: ISRO

Acknowledgement: Dr. VinothKumar S, DD/CHO, Madurai Corporation, TamilNadu Public Health Department, India

**Abstract**

This paper focuses on studying the impact of climate data and vector larval indices on dengue outbreak. After a comparative study of the various LSTM models, Bidirectional Stacked LSTM network is selected to analyze the time series climate data and health data collected for the state of Tamil Nadu (India), for the period 2014 to 2020. Prediction accuracy of the model is significantly improved by including the mosquito larval index, an indication of VBD control measure.

**Introduction**

Dengue Fever (DF), an outbreak prone viral infection is transmitted by Aedes mosquitoes, which is mostly found in tropical and sub-tropical climatic regions. The infection can result in Dengue Haemorrhagic Fever (DHF), also known as severe dengue which can be fatal. DF and DHF are caused by four serotypes of dengue virus (DENV). Till date there is no specific treatment for DF and DHF. Dengue prevention and effective management of control measures are necessary to reduce the catastrophic effects of dengue outbreaks.

With dengue being identified as one of the four major global health threats by the World Health Organization (WHO), India records approximately 33 million dengue cases each year, sharing one-third of the world's total dengue cases [1]. Implementation of advanced early warning systems and disease control measures is the need of the hour to mitigate dengue fatalities. Artificial intelligence-based solutions will go a long way in establishing active dengue surveillance needed to combat the disease outbreak [2].

A review of the literature on epidemiological study of dengue spread reveals the strong association between climate parameters and dengue transmission [3][4]. In addition to the meteorological data, socio-economic factors such as urbanization and population density also contribute for dengue dynamics [5][6][7]. Numerous other studies have considered including mosquito infection rate and vector surveillance data to improve the morbidity rate prediction further [8][9]. Novel approaches reported in the literature suggest the use of social media data such as Twitter for prediction with high spatial resolution [10] and remotely sensed data from satellites for modelling oviposition activity [11]. These provide us the motivation to exploit every available data to advance the research to the next level. Dengue dynamics has been modelled in the literature using various techniques ranging from classical methods such as AR, MA, ARMA, ARIMA and SARIMA to machine learning models such as SVM, Naïve Bayes, bagging and boosting algorithms and neural networks [12][13][14][15][16][17]. Classical methods are not without their limitations - they cannot learn non-linear relationships, not suitable for multivariate-forecasting and not robust to noise. In case of machine learning models, feature extraction is required to be done manually. Hence, there is a need to develop better models to improve the prediction accuracy [18][19].

The objective of this study is to use a deep learning model appropriate for processing time series data such as Long Short-Term Memory network, to evaluate the significance of climate parameters and vector larval indices on dengue outbreak.

**Dataset**

**Climate data**

Meteorological data (Temperature and Relative humidity) from Indian Government official weather portal and the weekly surface rainfall data from IMD, Pune are collected for the period 2014 to 2020. The climate readings are available for 26 districts in Tamilnadu.

**Health data**

Data pertaining to dengue incidence, mosquito larval index and control measures are collected from Tamilnadu public health department. The Domestic Breeding Checkers (DBC) are deployed to inspect the houses in each district and depending on the breeding count, each house is classified as belonging to one of the following three larval index categories - 0-5%, 5-10% and >10%. This results in three different counts for each district. Mosquito Larval Index is available for the period 2018 to 2020. A total of 2184 records are available which is split into train and test dataset in the ratio 85:15.

**Methodology**

**Data Preparation**

Daily climate data such as temperature and relative humidity collected for each district in Tamilnadu are processed to generate the monthly average. In order to feed mosquito larval index data as input to the network model, the three counts are combined by calculating the weighted average of mosquito index for each district as follows:

Weighted Average = (no. of houses with 0-5% * 1) + (no. of houses with 5-10% * 2) + (no. of house with >10% *3) / (Total Weight).

Weights are assigned proportionate to the larval index percentage such as 1 for 0-5%, 2 for 5-10% and 3 or >10%. This is for the reason that a higher percentage of larval count contributes more to the dengue incidence.

**Missing Data Imputation**

Mosquito larval index data has a number of missing values which the co-training algorithm imputes by modelling it as the target variable. Co-training is a semi-supervised regression algorithm called COREG. Two regressors, each with a sufficient view from the labelled dataset, are refined with the help of unlabelled examples that are labelled based on the highest level of confidence [19].

**Long Short-Term Memory Network**

LSTM is the deep learning model for time series prediction, which better learns the mapping of the underlying multivariate time series dataset. After hyper parameter tuning, the LSTM model is designed with 4 hidden layers and 3000 epochs. The input is formatted in a 3D tensor with 3 timesteps (One time step is one point of observation in a time series sample), that includes climate parameters and dengue incidence of the preceding two months and climate parameters alone for the current month.

The model learning process is improved by regularizing weights and a non-linear activation function, Relu is used. A dropout of 0.2 is used to reduce over-fitting and improve the generalization of the network model.

Two network model designs are experimented. In the first variant, only the climate parameters are used as the predictors. In the second variant, mosquito larval index is also included as an additional predictor, along with the climate parameters. Climate parameters, mosquito larval index of the preceding two months and the current month and dengue incidence of the preceding two months are given as input to the network model to predict the dengue incidence for the current month.

Error is estimated by using Mean Square Error (MSE), the average of the squared difference between the actual and the predicted target value where $\tilde{Y}_i$, is the predicted value of the i$^{th}$ test case.

$$MSE = \frac{1}{N}\sum_{i=0}^{N}(Y - \tilde{Y}_i)^2$$

**Model results and discussion**

Prediction accuracy of the variant II model is better than that of the variant I network model. Table 1 shows the prediction results of variant I which is trained with only the climate parameters. Table 2 shows the prediction results of variant II trained with both the climate parameters and the mosquito larval index.

Table 1. Dengue incidence prediction for 8 districts in Tamilnadu with dataset containing only climate parameters from 2018 – 2019

| Month | Dindigul | Madurai | Vellore | Trichy | Salem | Tirunelveli | Virudhunagar | Theni |
|---|---|---|---|---|---|---|---|---|
| Jan | 8 | 12 | 2 | 11 | 7 | 35 | 17 | 14 |
| Feb | 6 | 22 | 2 | 15 | 0 | 15 | 6 | 16 |
| Mar | 4 | 8 | 0 | 21 | 3 | 6 | 1 | 11 |
| Apr | 8 | 33 | 5 | 5 | 6 | 0 | 4 | 2 |
| May | 3 | 2 | 2 | 3 | 2 | 0 | 2 | 5 |
| Jun | 4 | 2 | 0 | 0 | 1 | 1 | 1 | 14 |
| Jul | 5 | 7 | 4 | 4 | 4 | 1 | 1 | 16 |
| Aug | 5 | 5 | 12 | 9 | 5 | 1 | 11 | 9 |
| Sep | 11 | 13 | 2 | 6 | 9 | -1 | 16 | 17 |
| Oct | 18 | 20 | 17 | 20 | 20 | 2 | 8 | 23 |
| Nov | 10 | 13 | 42 | 10 | 13 | 0 | 22 | 21 |
| Dec | 16 | 22 | 15 | 21 | 34 | 13 | 15 | 14 |
| Predicted Count | 97 | 161 | 104 | 123 | 103 | 72 | 105 | 161 |
| Actual Count | 49 | 279 | 19 | 157 | 93 | 74 | 41 | 149 |

Table 2. Dengue incidence prediction for 8 districts in Tamilnadu with dataset containing both Climate parameters and mosquito larval index from 2018 – 2019

| Month | Dindigul | Madurai | Vellore | Trichy | Salem | Tirunelveli | Virudhunagar | Theni |
|---|---|---|---|---|---|---|---|---|
| Jan | 11 | 55 | 3 | 51 | 7 | 34 | 5 | 23 |
| Feb | 4 | 44 | 2 | 27 | 1 | 17 | 5 | 24 |
| Mar | 4 | 31 | 1 | 11 | 2 | 5 | 4 | 12 |
| Apr | 4 | 7 | 1 | 8 | 4 | 2 | 2 | 5 |
| May | 1 | 1 | 1 | 2 | 1 | 1 | 2 | 6 |
| Jun | 2 | 4 | 0 | 2 | 1 | 1 | 2 | 2 |
| Jul | 3 | 5 | 4 | 5 | 7 | 2 | 2 | 2 |
| Aug | 1 | 2 | 4 | 7 | 7 | 1 | 4 | 3 |
| Sep | 4 | 4 | 6 | 10 | 19 | 2 | 2 | 10 |
| Oct | 3 | 20 | 6 | 6 | 7 | 1 | 4 | 31 |
| Nov | 7 | 18 | 2 | 12 | 17 | 4 | 12 | 10 |
| Dec | 6 | 16 | 3 | 10 | 25 | 1 | 15 | 6 |
| Predicted Count | 51 | 207 | 34 | 151 | 99 | 73 | 58 | 134 |
| Actual Count | 49 | 279 | 19 | 157 | 93 | 74 | 41 | 149 |

The time step in a time series is a single observation of a feature. To better understand the influence of time lag of the predictors in dengue incidence, the LSTM model is tested with four different time steps such as t = 2, 3, 4, 5 out of which the time step of 3 has the least mean square error for both the validation and test dataset (shown in Table 3).

Table 3. Mean Square error comparison for different time steps

| Time step | Mean Square Error (MSE) | |
|---|---|---|
| | Validation data | Test data |
| t = 2 | 9.39648 | 69.84859 |
| t = 3 | 5.63022 | 52.6196 |
| t = 4 | 5.89458 | 82.08041 |
| t = 5 | 31.2016 | 82.1811 |

In order to study the impact of the individual climate parameters on dengue prediction, the LSTM model is trained with a single climate parameter. The mean squared error of the model for both validation data and test data are listed in Table 4. The model trained with all the climate parameters performs well for both validation and test data.

Table 4. Mean Square error comparison for individual predictors

| Predictor | Mean Square Error (MSE) | |
|---|---|---|
| | Validation data | Test data |
| Temperature | 37.111 | 144.3402 |
| Rainfall | 17.44929 | 161.4387 |
| Relative Humidity | 30.7778 | 100.033 |
| All three parameters | 5.63022 | 52.6196 |

**Stacked LSTM:** In stacked LSTM, the model is built with multiple LSTM layers. The hidden LSTM layers make the model deeper.

**Bidirectional LSTM:** In Bidirectional LSTM, the model learns from the input sequence as well as the reverse sequence. Table 5 records the results of different types of LSTM models with the bidirectional stacked LSTM model yielding the best results.

Table 5. Mean Square error for different LSTM Models

| Model | Mean Square Error (MSE) | |
|---|---|---|
| | Validation data | Test data |
| LSTM | 5.63022 | 52.6196 |
| Stacked LSTM | 45.34371 | 70.89695 |
| Bidirectional LSTM | 11.4956 | 103.0428 |
| Bidirectional Stacked LSTM | 10.12985 | 67.29691 |

**Conclusion**

Bidirectional Stacked LSTM, a recurrent neural network has proved to make dengue predictions with good accuracy, when trained with the available climate data and mosquito larval index with a lead time of 3 months. With these promising results, the study can be extended further to include vegetation indices (Enhanced Vegetation Index, EVI and Normalized Difference Vegetation Index, NDVI) and human mobility patterns. In India, data digitization in the public health department has gained momentum only in the recent years that best explains the reason for inadequate mosquito larval index data. Still, transfer learning can be applied to achieve optimal performance even with a small dataset. The model can be leveraged to strengthen dengue surveillance and early warning systems across the country for the timely implementation of effective disease control measures.